\documentclass[11pt,a4paper]{article}
\usepackage[hyperref]{acl2020}
\usepackage{times}
\usepackage{latexsym}

\usepackage{microtype}

\aclfinalcopy 

\usepackage{xspace}
\usepackage{graphicx}

\usepackage{xcolor}
\usepackage{multirow}

\newcommand\gpttune{\textsc{GPT-Tuned}\xspace}
\newcommand\ptgen{\textsc{PtGen}\xspace}
\newcommand\tconv{\textsc{TConvS2S}\xspace}
\newcommand\tencdec{\textsc{TranS2S}\xspace}
\newcommand\bencdec{\textsc{BertS2S}\xspace}
\newcommand\gold{\textsc{Gold}\xspace}
\newcommand\xsum{\textsc{XSum}}

\definecolor{forestgreen}{HTML}{009B55}
\definecolor{sepia}{HTML}{671800}
\definecolor{midnightblue}{HTML}{006795}
\definecolor{orangered}{HTML}{ED135A}

\title{On Faithfulness and Factuality in Abstractive Summarization}

\author{Joshua Maynez\thanks{\quad The first two authors contributed equally.} \quad Shashi Narayan$^\ast$ \quad Bernd Bohnet \quad Ryan McDonald \\
Google Research \\
  {\small \tt \{joshuahm,shashinarayan,bohnetbd,ryanmcd\}@google.com}
}

\date{}

\begin{document}
\maketitle
\begin{abstract}
It is well known that the standard likelihood training and approximate decoding objectives in neural text generation models lead to less human-like responses for open-ended tasks such as language modeling and story generation. In this paper we have analyzed limitations of these models for abstractive document summarization and found that these models are highly prone to hallucinate content that is  unfaithful to the input document.
We conducted a large scale human evaluation of several neural abstractive summarization systems
to better understand the types of hallucinations they produce. Our human annotators found substantial amounts of hallucinated content in all model generated summaries. However, our analysis does show that pretrained models are better summarizers not only in terms of raw metrics, i.e., ROUGE, but also in generating faithful and factual summaries as evaluated by humans. Furthermore, we show that textual entailment measures better correlate with faithfulness than standard metrics, potentially leading the way to automatic evaluation metrics as well as training and decoding criteria.\footnote{Our human annotated summaries for faithfulness and factuality will be released at \href{https://github.com/google-research-datasets/xsum\_hallucination\_annotations}{https://github.com/google-research-datasets/xsum\_hallucination\_annotations}.}
\end{abstract}

\section{Introduction}

Current state of the art conditional text generation models accomplish a high level of fluency and coherence, mostly thanks to advances in sequence-to-sequence architectures with attention and copy \cite{Sutskever2014Sequence,bahdanau_2014,Gu2016Incorporating}, fully attention-based Transformer architectures \cite{transformer,transformetxl} and more recently pretrained language modeling for natural language understanding \cite{bert,gpt,xlnet_arxiv19,roberta}. There has been a growing interest in understanding how  maximum likelihood training and approximate beam-search decoding in these models lead to 
less human-like text in \textit{open-ended text generation} such as language modeling and story generation \cite{holtzman-arxiv19,welleck-arxiv19,see2019massively}.
In this paper we investigate 
how these models are prone to generate hallucinated text in \textit{conditional text generation}, specifically, extreme abstractive document summarization \cite{narayan18:xsum}.

\begin{figure*}[t!]
  \center{\footnotesize 
    \begin{tabular}{l p{10.6cm} l}
    \hline 
      
    
    \textbf{\gold} & \multicolumn{2}{p{12.5cm}}{\textcolor{midnightblue}{Zac Goldsmith} will \textcolor{midnightblue}{contest} the \textcolor{orangered}{2016} \textcolor{midnightblue}{London mayoral election} for the \textcolor{midnightblue}{Conservatives}, it has been announced.} \\ \hline

    \multicolumn{3}{p{15.5cm}}{\textbf{\textsc{Document: }}  The Richmond Park and North Kingston MP said \textbf{he was "honoured" after winning} 70\% of the 9,227 votes cast using an online primary system. } \\ 

    \multicolumn{3}{p{15.5cm}}{He beat London Assembly Member Andrew Boff, MEP Syed Kamall and London's deputy mayor for crime and policing Stephen Greenhalgh.} \\

    \multicolumn{3}{p{15.5cm}}{\textbf{Mr Goldsmith}'s main rival is likely to be \textbf{Labour's Sadiq Khan}. \textit{(2 sentences with 59 words are abbreviated here.)}}\\

    \multicolumn{3}{p{15.5cm}}{\textbf{Mr Goldsmith}, who was \textbf{the favourite for the Tory nomination}, balloted his constituents earlier this year to seek permission to stand. } \\ 
    
    \multicolumn{3}{p{15.5cm}}{At the very point of \textbf{his entry into the race for London mayor}, \textbf{Zac Goldsmith}'s decision revealed two big characteristics. \textit{(5 sentences with 108 words are abbreviated here.)} } \\
    
    
    \multicolumn{3}{p{15.5cm}}{\textbf{Mr Goldsmith} - who first entered Parliament in 2010 - told the BBC's Daily Politics that he hoped his environmental record would appeal to Green and Lib Dem voters and he also hoped to "reach out" to \textbf{UKIP} supporters frustrated with politics as usual and the UK's relationship with the EU.} \\
    
    \multicolumn{3}{p{15.5cm}}{\textbf{Zac Goldsmith} Born in 1975, educated at Eton and the Cambridge Centre for Sixth-form Studies \textit{(5 sentences with 76 words are abbreviated here.)}} \\
    
    
    \multicolumn{3}{p{15.5cm}}{\textbf{Mr Goldsmith}, who has confirmed he would stand down from Parliament if he became mayor, triggering a by-election, said he wanted to build on \textbf{current mayor Boris Johnson}'s achievements. \textit{(3 sentences with 117 words are abbreviated here.)}} \\




    \multicolumn{3}{p{15.5cm}}{Both \textbf{Mr Khan and Mr Goldsmith} oppose a new runway at Heathrow airport, a fact described by the British Chambers of Commerce as "depressing". \textit{(1 sentences with 31 words is abbreviated here.)}} \\


    \multicolumn{3}{p{15.5cm}}{\textbf{Current mayor Boris Johnson} will step down next year after two terms in office. He is also currently the MP for Uxbridge and South Ruislip, having been returned to Parliament in May.} \\

    \multicolumn{3}{p{15.5cm}}{Some \textbf{Conservatives} have called for an inquiry into \textbf{the mayoral election process} after only 9,227 people voted - compared with a 87,884 turnout for the Labour contest. \textit{(4 sentences with 121 words are abbreviated here.)}} \\




    
    \hline
    
    \textbf{\ptgen} & \textcolor{orangered}{UKIP leader Nigel Goldsmith} has been \textcolor{orangered}{elected} as the \textcolor{orangered}{new mayor of London} to \textcolor{orangered}{elect a new Conservative MP}. &[45.7, 6.1, 28.6]  \\

    \textbf{\tconv} & \textcolor{orangered}{Former London mayoral candidate} \textcolor{midnightblue}{Zac Goldsmith} has been \textcolor{midnightblue}{chosen to stand} in \textcolor{midnightblue}{the London mayoral election}. & [50.0, 26.7, 37.5] \\
    
    \textbf{\tencdec} & \textcolor{orangered}{Former London mayor} \textcolor{midnightblue}{Sadiq Khan} has been \textcolor{midnightblue}{chosen as the candidate} to be \textcolor{midnightblue}{the next mayor of London}. & [35.3, 12.5, 23.5] \\
    
    \textbf{\gpttune} & \textcolor{orangered}{Conservative MP Zac Goldwin}'s bid to become \textcolor{orangered}{Labour's candidate} in the \textcolor{orangered}{2016} \textcolor{midnightblue}{London mayoral election}. & [42.4, 25.8, 36.4] \\
    \textbf{\bencdec} & \textcolor{midnightblue}{Zac Goldsmith} has been \textcolor{midnightblue}{chosen to contest the London mayoral election}. & [66.7, 40.0, 51.9] \\
    
    \hline
    \end{tabular}     
  }
  \caption{Hallucinations in extreme document summarization: the abbreviated article, its gold summary and the abstractive model generated summaries (\ptgen, \citeauthor{see-acl17} \citeyear{see-acl17};  \tconv, \citeauthor{narayan18:xsum} \citeyear{narayan18:xsum}; 
  and, \gpttune, \tencdec and \bencdec, \citeauthor{berts2s} \citeyear{berts2s}) for a news article from the extreme summarization dataset \cite{narayan18:xsum}. The dataset and the abstractive models are described in Section~\ref{sec:xsum} and~\ref{sec:abssys}. We also present the [\textsc{rouge-1}, \textsc{rouge-2}, \textsc{rouge-l}] F$_1$ scores relative to the reference gold summary. Words in \textcolor{orangered}{red} correspond to hallucinated information whilst words in \textcolor{midnightblue}{blue} correspond to faithful information.}\label{fig:erroranalysis-xsum}
  \vspace{-0.6cm}
\end{figure*}

Document summarization --- the task of producing a shorter version of a document while preserving its information content \cite{mani2001automatic,Nenkova:McKeown:2011} --- requires models to generate text that is not only human-like but also faithful and/or factual given the document. The example in Figure~\ref{fig:erroranalysis-xsum} illustrates that the faithfulness and factuality are yet to be conquered by conditional text generators. The article
describes an event of ``\textit{Conservative MP Zac Smith winning the primary for 2016 London mayoral election}'', but summaries often forge entities (e.g., ``Nigel Goldsmith'' or ``Zac Goldwin'') or information (e.g., ``UKIP leader Nigel Goldsmith'', ``Nigel Goldsmith winning the mayoral election'', ``Sadiq Khan being the former London mayor'' or ``Zac Goldwin being the Labour's candidate'') that are not supported by the document or are factually wrong. Interestingly, all summaries are topical and fluent, and perform well in terms of ROUGE scores \cite{rouge}. 

We conducted a large-scale human evaluation of hallucinated content in systems that use Recurrent Neural Network (RNN)  \cite{see-acl17}, Convolutional Neural Network (CNN) \cite{narayan18:xsum}, and Transformers \cite{gpt2,berts2s}, as well as human written summaries for the recently introduced e\textsc{X}treme \textsc{Sum}marization task \citep[\xsum,][]{narayan18:xsum}. We seek to answer the following questions: (i) How frequently do abstractive summarizers hallucinate content?; (ii) Do models  hallucinate by manipulating the information present in the input document (\textit{intrinsic hallucinations}) or by adding information not directly inferable from the input document (\textit{extrinsic hallucinations})?;  (iii) How much hallucinated content is \emph{factual}, even when unfaithful?; and (iv) Are there automatic means of measuring these hallucinations?

Our main conclusions are as follows: First, intrinsic and extrinsic hallucinations
happen frequently  -- in more than 70\% of single-sentence summaries. Second, the majority of hallucinations are extrinsic, which potentially could be valid abstractions that use background knowledge. However, our study found that over 90\% of extrinsic hallucinations were erroneous.
Thus, hallucinations happen in most summaries and the majority of these are neither faithful nor factual. Third, models initialized with pretrained parameters perform best both on automatic metrics and human judgments of faithfulness/factuality. Furthermore, they have the highest percentage of extrinsic hallucinations that are factual. This suggests that while some studies argue that large-scale pretrained models are merely better at learning data-specific regularities~\cite{niven2019}, at least on in-domain summarization the gains in automatic metrics are realized in observable differences by humans. Fourth, ROUGE~\cite{rouge} and BERTScore \cite{bertscore} correlates less with faithfulness/factuality than metrics derived from automatic semantic inference systems, specifically the degree to which a summary is entailed by the source document. This presents an opportunity for improved automatic evaluation measures as well as model training and decoding objectives. We show preliminary experiments in this direction.

\section{Hallucinations in Summarization}
\label{sec:defineHal}

Open-ended generation --- the task of generating text that forms a natural continuation from the input text --- requires the model to hallucinate text; hence the focus has been to ensure that the model learns to generate text that is more human-like (i.e., less repetitive or dull with more content-related words) \cite{holtzman-arxiv19,welleck-arxiv19,see2019massively}.
In contrast, tasks such as document summarization \cite{Nenkova:McKeown:2011,see-acl17,Paulus2018Deep} and data-to-text generation \cite{lebret2016neural,wiseman-etal-2017-challenges} which are not open-ended, require models to be factual and/or faithful to the source text.

Despite recent improvements in conditional text generation, most summarization systems are trained to maximize the log-likelihood of the reference summary at the word-level, which does not necessarily reward models for being faithful. Moreover, models are usually agnostic to the noises or artifacts of the training data, such as reference divergence, making them vulnerable to hallucinations \cite{Kryscinski2019NeuralTS,wiseman-etal-2017-challenges,dhingra-etal-2019-handling}. Thus, models can generate texts that are not consistent with the input, yet would likely have reasonable model log-likelihood.

\subsection{Intrinsic and Extrinsic Hallucinations}
\label{subsec:haltypes}

Given a document $D$ and its abstractive summary $S$, we try to identify all hallucinations in $S$ with respect to the content of $D$, regardless of the quality of the summary. In this work, we define a summary as being hallucinated if it has 
a span(s) $w_i\dots w_{i+j}$, $j \geq i$, that is not supported by the input document. 
To distinguish hallucinations further in the context of a document and a summary, we categorize hallucinations by the information source as \textit{intrinsic} and \textit{extrinsic} hallucinations. Note, paraphrases or any information that can be inferred from the document are not categorized as hallucinations. 

\textbf{Intrinsic hallucinations} are consequences of synthesizing content using the information present in the input document. For example, in Figure~\ref{fig:erroranalysis-xsum}, ``Former London mayoral candidate'' in the \tconv abstract and ``Former London mayor'' in the \tencdec abstract are hallucinations of intrinsic nature; both use terms or concepts from the document but misrepresent information from the document, making them unfaithful to the document. The article does not confirm if ``Zac Goldsmith'' was a ``Former London mayoral candidate'' or if ``Sadiq Khan'' was a ``Former London mayor''. One may suspect that a model with poor input document representation will fail to do document level inference, often required for abstraction, and will be vulnerable to such errors. 

\textbf{Extrinsic hallucinations} are model generations that ignore the source material altogether. For example, in Figure~\ref{fig:erroranalysis-xsum}, ``Nigel'' in the \ptgen abstract
and ``2016'' in both \gold and \gpttune are extrinsic hallucinations; these terms are not introduced in the document. A model with a poorly-informed decoder and that is agnostic to the divergence issue between the source and target texts \cite{wiseman-etal-2017-challenges,dhingra-etal-2019-handling}, will function more as an open-ended language model and will be prone to extrinsic hallucinations. 

\subsection{Factual Hallucinations in Summarization}
\label{subsec:factualhal}

A summary $S$ of a document $D$ contains a \emph{factual hallucination} if it contains information not found in $D$ that is factually correct. Factual hallucinations may be composed of intrinsic hallucinations or extrinsic hallucinations.

By definition, abstractive summaries are written to preserve the salient information in the input document, but they are  expressed in the words of the summary author as opposed to the input document author \cite{Nenkova:McKeown:2011}. As such, it is natural to construct summaries that integrate with the author's background knowledge \cite{dijk1978,brown1983}. Such knowledge integration can also be desirable in real world applications. For instance, an engaging sports report will reflect an understanding of the game to provide color and context. Another example is audience-targeted summarization where a good summary will reflect understanding of both the article domain and the desired audience. Nonetheless, there is no consensus in the research community if the summary should be faithful (without any hallucinations) to the input document or if there is tolerance for factual hallucinations. 

Recent deep learning approaches to abstractive summarization naturally learn to integrate knowledge from the training data while generating an abstractive summary for a document \cite{see-acl17,gehrmann2018bottom}. More advanced pretrained text generators \cite{gpt,gpt2,unilm_arxiv19,mass_icml19,khandelwal_2019,berts2s} are even better at capturing world knowledge as they are informed by a vast amount of background text. This can be observed in the example shown in Figure~\ref{fig:erroranalysis-xsum} as the input document does not mention that the discussed ``London mayoral election'' is from ``2016''; but the abstract generated by the pretrained text generator \gpttune correctly predicts this information similar to the human-authored abstract.\footnote{Despite the correct extrinsic hallucination (``2016 ''), the \gpttune abstract overall is still not factual due to the incorrect extrinsic hallucination in ``Conservative MP Zac Goldwin.'' There is no Conservative MP named Zac Goldwin.}

In this paper we stand in favour of the assertion that abstractive systems may integrate with the background knowledge to generate rich and meaningful summaries. More concretely, ``\textit{hallucinations in summarization are acceptable if they lead to better summaries that are factual with respect to the document and the associated background knowledge}.'' This assumption also allows us to assess the capability of recent neural models to integrate with the background knowledge to generate factual abstracts (see Section~\ref{subsec:humaneval-b}).

\section{Extreme Document Summarization}
\label{sec:xsum}

We focus on the recently introduced extreme summarization
dataset \citep[\xsum,][]{narayan18:xsum}\footnote{\url{https://github.com/EdinburghNLP/XSum}} which comprises 226,711 British Broadcasting Corporation (BBC) articles paired with their single-sentence summaries, provided by the journalists writing the articles. The dataset is split into three subsets: training (90\%, 204,045), validation (5\%, 11,332), and test (5\%, 11,334) sets. All models in \S\ref{sec:abssys} trained to generate abstractive summaries are trained and evaluated using this standard split, provided by the authors.

We choose to focus our study on extreme summarization for the following reasons: First, this task aims to create a single-sentence summary of a news article; these shorter summaries are relatively easier to annotate and analyze than longer summaries such as story highlights from the CNN/Dailymail dataset \cite{hermann-nips15} or abstracts from the NY Times \cite{nytcorpus} or the WikiSum \cite{wikisum} dataset. Secondly, the gold summary in the extreme summarization dataset is an introductory sentence prefacing each article. By virtue of this property, the extreme summarization task is not amenable to extractive strategies and requires an abstractive modeling approach. Hence, it provides us a better benchmark to assess abstractive models' abilities to produce abstractions which are faithful and factual. Finally, since we conclude that hallucination is a problem on this dataset, then we can safely conclude it is a problem for summarization datasets with longer summaries, as modeling longer-distance dependencies and discourse structures make the task harder.

\section{Abstractive Summaries}
\label{sec:abssys}

We evaluate summaries from RNN, CNN and Transformer-based state-of-the-art abstractive summarization methods and the reference human written summaries. See the Appendix for hyperparameter and decoding details for all models.

\paragraph{Human Written Reference Summaries.}

The single-sentence summaries contained in the extreme summarization dataset (\xsum) are also evaluated as part of this study. These summaries were written by journalists as introductions to the news articles they precede. These summaries, therefore, often have true additional information not found in the document. Such divergence issue between source and target is not uncommon in conditional text generation \cite{Kryscinski2019NeuralTS,wiseman-etal-2017-challenges,dhingra-etal-2019-handling}.

\paragraph{RNN-based Seq2Seq.} 

We use the Pointer-Generator model (\textbf{\ptgen}) introduced by \newcite{see-acl17}, an RNN-based attention-based sequence to sequence model which not only generates from the target vocabulary but can copy words from the source text.

\paragraph{Topic-aware Convolutional Seq2Seq.} 

The Topic-aware Convolutional Sequence to Sequence model (\textbf{\tconv}) introduced by \newcite{narayan18:xsum} is an abstractive system which is conditioned on the article's topics and based entirely on Convolutional Neural Networks \cite{convseq2seq}. \tconv is better suited for extreme summarization, as convolution layers capture long-range dependencies between words in the document more effectively than RNNs. Simultaneously, the convolutional encoder associates each word with a topic vector, capturing whether it is representative of the document's content.

\paragraph{Transformer-based Abstractive Methods.} 

We experiment with three Transformer-based model variants, all of which have 12 layers, a hidden size of 768, filter size of 3072, and 12 attention heads.

\noindent \textbf{\gpttune}: \newcite{gpt2} proposed Transformer-based Generative Pre-Trained (GPT) language models that can generate high quality text in open-ended generation setups. 
The proposed decoder-only architecture for language modeling can be easily adapted to abstractive summarization where the model first sees the document and, given a prompt, such as TL;DR;, generates its summary. Our \gpttune is warm-started with a publicly available GPT checkpoint \cite{gpt2}, but fine-tuned with supervised training on the extreme summarization dataset.

\noindent \textbf{\tencdec} and \textbf{\bencdec}: \tencdec and \bencdec are sequence to sequence models where both encoder and decoder are composed of Transformer layers \cite{transformer,berts2s}. 
All weights in \tencdec are randomly initialized, but in \bencdec, both encoder and decoder are initialized with the BERT-Base checkpoints \cite{bert}, with parameter sharing between the encoder and decoder, following \newcite{berts2s}. The only variable that is initialized randomly is the encoder-decoder attention in \bencdec. Both models are then trained on the extreme summarization dataset.

\section{Experiments and Results}
\label{sec:exp-result}

The main focus of this work is not to propose a solution to hallucination related issues, but to achieve a better understanding of hallucinations in abstractive summarization through their human assessment. We randomly sampled 500 articles from the test set to facilitate our study. Using the full test set was unfeasible given its size and the cost of human judgments. We have trained annotators (fluent in English) specifically for our assessment. Our annotators went through two pilot studies to have a better understanding of intrinsic and extrinsic hallucinations, and factuality of summaries. Documents used in the pilot studies were not used in the final annotation. We also report on ROUGE \cite{rouge} scores, BERTScore \cite{bertscore} and semantic inference metric such as textual entailment \cite{Pasunuru-multireward18,welleck-etal-2019-dialogue,falke-etal-2019-ranking,Kryscinski2019EvaluatingTF} and question answering \cite{arumae-liu-2019-guiding,Wang2020AskingAA}.

\begin{table}[t!]
\centering
\footnotesize
\begin{tabular}{l|cccc}
\hline
\multirow{2}{*}{\textbf{Models}} & \multicolumn{4}{|c}{\textbf{Human Eval Test Set}} \\
& \textsc{r1} & \textsc{r2} & \textsc{rl} & BERTScore \\ \hline
\ptgen & 30.01 & 9.38 & 23.76 & 74.30  \\ 
\tconv & 30.89 & 11.47 & 25.80 & 75.23 \\
\tencdec & 32.28 & 11.66 & 24.65 & 75.69\\ \hline
\gpttune & 21.82 & 4.72 & 16.28 & 70.35  \\ 
\bencdec & \textbf{38.42}& \textbf{16.96} & \textbf{31.27} & \textbf{78.85} \\ \hline
\end{tabular}
\caption{ROUGE and BERTScore F$_1$ scores for non-pretrained (the top block) and pretrained (the bottom block) models reported on the XSum dataset. These results are on the sampled human evaluation (500 items) dataset. The best results are \textbf{boldfaced}.}
\label{tab:rouge}
\vspace{-0.5cm}
\end{table}

\subsection{Automatic Evaluation of Summaries}
\label{subsec:rouge}

ROUGE \cite{rouge} provides a means to quickly assess a model's ability to generate summaries closer to human authored summaries. We report on ROUGE-1 and ROUGE-2 for informativeness and ROUGE-L, for fluency. 
Like ROUGE, BERTScore \cite{bertscore} computes a similarity score for each token in the candidate summary with each token in the reference summary. However, instead of exact matches, it computes token similarity using contextual embeddings. Results are presented in Table~\ref{tab:rouge}.

For both cases, the pretrained encoder-decoder architecture \bencdec performed far superior to any other randomly initialized models, such as \ptgen, \tconv and \tencdec, and the decoder-only architecture \gpttune. 
The differences between \ptgen, \tconv and \tencdec are not significant; all other differences are significant.\footnote{Pairwise comparisons between all models using a one-way ANOVA with post-hoc Tukey HSD tests; $p < 0.01$.}

ROUGE and BERTScore are indicators of informativeness of summaries but they are not sufficient metrics to assess the overall quality of summaries.
This becomes evident from our human assessments in the following sections where we employ human annotators to evaluate summaries generated with \ptgen, \tconv, \tencdec and \bencdec, and the human authored summaries. We excluded \gpttune abstracts from our study after their poor performance on the automatic measures.

\subsection{Assessment of Hallucinations}
\label{subsec:humaneval-a}

\begin{figure}[t!]
  \begin{center}
  \includegraphics[width=0.98\linewidth]{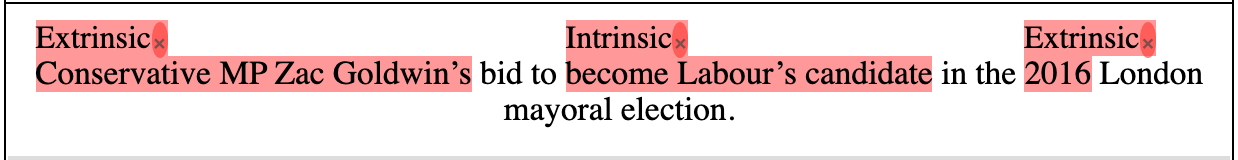}
  \vspace{-0.2cm}
  \caption{Human assessment of a system generated summary for the article in Figure~\ref{fig:erroranalysis-xsum}. The annotation user interface is shown as it was shown to raters.}
  \label{fig:example-annotation}
  \end{center}
  \vspace{-0.6cm}
\end{figure}

In this assessment, human annotators were presented an article and a single-sentence summary for the article. They were stringently told to only assess the hallucinations in the summary and to not confuse their assessment with the quality of the summary. 
For summaries containing hallucinations, annotators were tasked with (i) identifying those text spans that were unfaithful to the article and (ii) for each text span, annotating whether the hallucination was intrinsic or extrinsic. We elicited judgments from three different annotators for each of 2500 (500x5) document-summary pairs. Figure~\ref{fig:example-annotation} shows an example assessment of a summary of an article from  Figure~\ref{fig:erroranalysis-xsum}. Results from the full assessment are shown in Table~\ref{tab:intexhal}, which shows the percentage of documents per system that were annotated as faithful or hallucinated (faithful = 100 - hallucinated). The Appendix provides inter-annotator agreement of hallucinations as well as hallucinated span characteristics.

\paragraph{Extrinsic Hallucination due to Divergence between Source and Target.}
Our results confirmed that the BBC gold summaries often have extrinsic hallucinations due to the dataset artifact that gold summaries are introductory sentences prefacing each article. It was not surprising that most models also had significant extrinsic hallucinations.

\paragraph{Intrinsic Hallucination is Also Common in Abstractive Summaries.} 

Gold summaries can also display intrinsic hallucinations. For example, a news article could describe an event related to ``Barack Obama'' and ``the office of the President of the United States'' without inferring that ``Obama is the President of the United States.'' A journalist with the knowledge of the event in the article could write a summary stating ``President Obama.'' 

However, the percentage of system summaries with intrinsic hallucination was much higher than in gold summaries (7.4\% vs others).
This phenomenon particularly revealed the models' tendency to misrepresent information in the document due to the lack of document-level understanding and inference. The copy mechanism in \ptgen is good at copying from the source (showing the least percentage of extrinsic hallucination of 63.3\%), but the mechanism lacks the inference capability and is prone to generate a summary that is not supported by the document (19.9\% intrinsic hallucination). \tencdec showed similar performance to \ptgen and ranked second worst. The \bencdec showed the least number of intrinsic hallucination (16.9\%) among all four abstractive systems. 

\begin{table}[t!]
\centering
\footnotesize
\begin{tabular}{l|ccc|cc}
\hline
\multirow{2}{*}{\textbf{Models}} & \multicolumn{3}{|c|}{\textbf{Hallucinated}} &  \multirow{2}{*}{\textbf{Faith.}} & \multirow{2}{*}{\textbf{+Fact.}}\\
& \textbf{I} & \textbf{E}  & \textbf{I $\cup$ E}  & \\ \hline
\ptgen & 19.9 & \textbf{63.3} & 75.3 & 24.7 & 27.3  \\ 
\tconv & 17.7 & 71.5 & 78.5 & 21.5 & 26.9  \\
\tencdec & 19.1 & 68.1 & 79.3 & 20.7 & 25.3 \\
\bencdec & 16.9 & 64.1 & \textbf{73.1} & \textbf{26.9} & \textbf{34.7}\\
\gold & \textbf{7.4} & 73.1 & 76.9 & 23.1 & --- \\ \hline
\end{tabular}
\caption{Intrinsic vs.\ Extrinsic Hallucinations. The numbers in ``Hallucinated'' columns show the percentage of summaries where at least one word was annotated by all three annotators as an intrinsic (I) or extrinsic (E) hallucination.
When a summary is not marked with any hallucination, it is ``faithful'' (100 - I$\cup$E), column ``Faith.''. The final ``+Fact.'' column shows the total percentage of faithful and/or factual summaries, which includes all faithful summaries plus the percentage of non-faithful summaries annotated by all three annotators as factual. Higher numbers for faithful/factual and lower numbers for hallucinations are \textbf{boldfaced}.}
\label{tab:intexhal}
\vspace{-0.5cm}
\end{table}

\paragraph{Pretraining Improves Faithfulness.}

Hallucinations do not result from the artifacts in the training data only, but also due to model shortcomings. The \ptgen model with the copy mechanism \cite{Gu2016Incorporating,see-acl17} had the lowest extrinsic hallucination (63.3\%), but \bencdec reported the highest number of faithful summaries.
It appears that \bencdec is overall most conservative among all four abstractive systems while getting closer to reference summaries in terms of ROUGE. 
The pretraining prepares BertS2S to be more aware of the domain of the document and less prone to language model vulnerabilities. Consequently, BertS2S is more confident in predicting tokens from the document than TranS2S, hence, improving faithfulness.

\subsection{Assessment of Factual Hallucinations.}
\label{subsec:humaneval-b}

Hallucinations are not necessarily erroneous. In our second human assessment, we measured to what extent this is the case. Our annotators were presented a single-sentence summary with hallucinations and were asked to assess whether it is true or false. 
To better explain the context of the summary, annotators were made available the source document as well as the external resources such as Wikipedia or Google Search.
The source document can be particularly important for generic summaries to better understand context.
External resources assisted the evaluators to validate grounded facts in public knowledge bases.

Annotators were expected to validate the summary by looking for supporting evidence for the information found on the summary. If information in the summary contradicts the document, then the summary is not factual. If supporting evidence is found for all the information, then the summary is factual. The document is not useful when the summary has information that is neither supported nor contradicted in the article.
For example, the summary in Figure~\ref{fig:example-annotation} mentions ``Conservative MP Zac Goldwin'' which can not be verified from the article in Figure~\ref{fig:erroranalysis-xsum}. Here, annotators could use Wikipedia or Google Search to check that there had not been a Conservative MP named Zac Goldwin who tried to change their party and become a Labour's candidate in the 2016 London mayoral election.

We dropped the human authored gold summaries from this evaluation; they were presumably factual. We also dropped the abstracts that were faithful to their input documents from the previous study. Finally, there were 1869 document-summary pairs where the summaries were marked with at least one intrinsic or extrinsic hallucination. We elicited judgments from three different annotators for each of them. Results from this assessment are also presented in Table~\ref{tab:intexhal} (see the column labelled ``+Fact.'') along with the hallucination assessment. 

\paragraph{Pretraining Helps Generating Factual Summaries.} 

In total, 34.7\% of the \bencdec abstracts were faithful (26.9\%) and/or factual (+7.8\%). This is 7.4\% absolute better than the next-best model (\ptgen). The number of unfaithful yet factual summaries for \bencdec, 7.8\%, was also the highest. In fact, for extrinsic hallucinations, even though \ptgen hallucinates less than \bencdec (63.3\% vs. 64.1\%), 6.6\% of \bencdec hallucinations were  factual, compared to 2.2\% of \ptgen.\footnote{See Appendix for full results.} Thus, if we consider factual hallucinations to be valid, this means that even for extrinsic cases, \bencdec hallucinates the least.

The superior performance of \bencdec is most likely due to its exposure to vast amount of text through pretraining, allowing it to integrate background knowledge with generation. Even so, over 90\% of \bencdec hallucinations are  erroneous.

Finally, we carried out pairwise comparisons between all models (using a one-way ANOVA with post-hoc Tukey HSD tests; $p < 0.01$). For intrinsic hallucinations (the second column in Table~\ref{tab:intexhal}), \gold is significantly different from all other systems. For extrinsic hallucinations (the third column in Table~\ref{tab:intexhal}), there were significant differences between \ptgen and \tconv, \ptgen and \gold, and, \bencdec and \gold. For factuality, the differences between \ptgen, \tconv, and \tencdec were insignificant.

\subsection{Automatic Measures for Hallucinations}
\label{subsec:nlumeasures}

Summaries are a proxy for their source documents under the assumption that they highlight the most important content. With this assumption, we further studied the extent to which the hallucinated content can be measured by semantic inference related measures, such as textual entailment and question answering.

\begin{table}[t!]
\centering
\footnotesize
\begin{tabular}{l|ccc|c}
\hline
\multirow{2}{*}{Models} & \multicolumn{3}{c|}{\textbf{Textual Entailment}} & \multirow{2}{*}{\textbf{QA}}\\ 
& \textbf{entail.} & \textbf{neut.} & \textbf{cont.} & \\ \hline
\ptgen & 38.4 & 34.4 & 27.2 & 20.2 \\ 
\tconv & 29.6 & 37.4 & 33.0 & 19.9 \\
\tencdec & 34.6 & 39.8 & 25.6 & 22.4 \\
\bencdec & \textbf{41.8} & 37.8 & 20.4 & \textbf{23.0} \\ 
\gold & 32.8 & \textbf{47.2} & \textbf{20.0} & 19.3 \\ \hline 
\end{tabular}
\caption{Textual entailment and question answering (QA)
based measures for summary evaluation. For entailment, we show the percentage of times a summary entails (entail.) the document, is neutral (neut.) to the document and contradicts (cont.) the document. For QA, we report the percentage of questions that were correctly answered by a system. The highest numbers for entail., neut. and QA, and the lowest number for cont. are boldfaced.}
\label{tab:proxy}
\vspace{-0.5cm}
\end{table}

\paragraph{Textual Entailment.} We trained an entailment classifier by finetuning a BERT-Large pretrained model \cite{bert} on the Multi-NLI dataset \cite{mnli}. We calculated the entailment probability score between the document and its abstractive summaries. Note that this entailment classifier is not optimal for the BBC article-summary pairs; the Multi-NLI dataset contains sentence-sentence pairs.

Ideally a summary should entail the document or perhaps be neutral to the document, but never contradict the document. As can be seen in Table~\ref{tab:proxy}, the \bencdec abstracts showed the least number of contradictions compared to other system-generated abstracts and was at par with the \gold summaries. Similar to the performance on extrinsic hallucination in Table~\ref{tab:intexhal}, the \tconv abstracts also displayed the highest number of contradictions. Interestingly, the \gold summaries are more neutral to their documents, whereas the \bencdec summaries are more entailed by their documents. This is probably due to the nature of the data and that journalists tend to add color and have a high number of extrinsic (but valid) hallucinations.

\paragraph{Question Answering.} QA frameworks have been used to assess or promote summary informativeness \cite{Narayan2018Ranking,arumae-liu-2019-guiding}.
We adapted the QA framework to assess hallucination in model generated summaries; a faithful model will generate a summary that only has information that is supported by its document. Under this assumption, any question answerable by the summary should also be answerable by the source.

Given an abstractive summary, we used the round-trip consistency method of \newcite{alberti-etal-2019-synthetic}, which combines question generation and answer extraction models to generate synthetic question-answer pairs. For the 500 document-summary pairs, we generated 731, 708, 720, 725 and 820 question-answer pairs for \ptgen, \tconv, \tencdec, \bencdec and \gold, respectively. Finally, we used a machine reading comprehension model to answer these questions using the document as context. As in \newcite{alberti-etal-2019-synthetic}, we trained all models: question generation, answer extraction and reading comprehension models; using a BERT-Base pretrained model \cite{bert} finetuned on the Natural Questions dataset \cite{nq}. 

\begin{figure}[t!]
    \footnotesize
    \begin{tabular}{p{1.4cm} p{5.6cm}}
    \hline
    \textbf{\ptgen} &
    Leeds United \textcolor{orangered}{fought back from 2-0 down to beat} Huddersfield town in the \textcolor{orangered}{first round of the EFL cup}. (\textbf{Q: }{\em What team did Leeds United beat in the first round of the EFL cup?}, \textbf{A: }{\em Huddersfield town})\\ 
    \hline
    \textbf{\tconv} &
    A coal mine in South Yorkshire has \textcolor{orangered}{collapsed as a result of the loss of a coal mine}. (\textbf{Q: }{\em What type of mine has collapsed?}, \textbf{A: }{\em Coal}) \\
    \hline
    \textbf{\tencdec} & Star Wars actor \textcolor{orangered}{James} Davis said \textcolor{orangered}{he was ``locked in a caravan''} and had his caravan stolen during a \textcolor{orangered}{break-in}. (\textbf{Q: }{\em Who said he was locked in a caravan?}, \textbf{A: }{\em Davis}) \\
    \hline
    \end{tabular}
    \vspace{-0.2cm}
   \caption{Sample of question-answer pairs generated from hallucinated summaries that are correctly answered by their source articles. 
    Highlighted \textcolor{orangered}{spans} in the summaries are marked as extrinsic or intrinsic hallucinations by our annotators.}
    \label{fig:question-answers}
    \vspace{-0.1cm}
\end{figure}

Similar to textual entailment results, the \bencdec abstracts were the most faithful to their source documents in terms of question answering. The \gold abstracts were the least accurate due to a high number of extrinsic hallucination in them.

\begin{table}[t!]
\centering
\footnotesize
\begin{tabular}{l | c  c}
\hline
\textbf{Metric} & \textbf{Faithful} & \textbf{Factual} \\ \hline
\textsc{rouge-1} & 0.197 & 0.125 \\
\textsc{rouge-2} & 0.162 & 0.095 \\
\textsc{rouge-l} & 0.162 & 0.113 \\ 
BERTScore & 0.190 & 0.116 \\ \hline
QA & 0.044 & 0.027 \\ 
Entailment & \textbf{0.431} & \textbf{0.264} \\
\hline
\end{tabular}
\vspace{-0.1cm}
\caption{Spearman's correlation coefficient ($|r_s|$) of different metrics with faithful and factual annotations.}
\label{tab:correl}
\vspace{-0.5cm}
\end{table}


\paragraph{Spearman's Correlation.} We estimate Spearman's correlation coefficients of different metrics with the faithful and factual human scores (see Table~\ref{tab:correl}). We found that the textual entailment scores are best correlated with both faithful (moderate, $0.40 \leq |r_s| \leq 0.59$) and factual (weak, $0.20 \leq |r_s| \leq 0.39$) human scores. Comparatively, \textsc{rouge}-based metrics and BERTScore have very weak correlation, our findings are consistent with the  recent studies \cite{fact-brainsum,Kryscinski2019NeuralTS,Wang2020AskingAA}. Surprisingly, the question answering scores showed a very weak correlation ($0.0 \leq |r_s| \leq 0.19$) with faithful and factual human scores. 
We hypothesize that this is due to a compounding of errors where (i) the question generator is used to generate questions from the systems' generated abstracts, instead of human-written text on which they were trained, (ii) the question generator is susceptible to generate questions  with hallucinated content when fed in with hallucinated summaries, and (iii) our assumption that a summary is faithful if the answers from the source and the summary match, is rather poor for extreme summarization. We demonstrate these issues in Figure~\ref{fig:question-answers}. 
Irrespective of questions with hallucinated content, our reading comprehension model can fortuitously answer them correctly from their source articles.
Better ways of generating questions \cite{Narayan2020QURIOUSQG} and measuring factual consistency may alleviate some of these issues \cite{Wang2020AskingAA}.

\begin{table}[t!]
\centering
\footnotesize
\begin{tabular}{l|ccc|cc}
\hline
\textbf{Models} & \textbf{\textsc{r1}} & \textbf{\textsc{r2}} & \textbf{\textsc{rl}} & \textbf{Faith.} & \textbf{+Fact.}\\\hline
\bencdec & \textbf{38.42} & \textbf{16.96} & \textbf{31.27} & 26.9 & 34.7\\
\textsc{entail} & 35.93 & 14.02 & 28.87 & 31.5 & 38.6 \\ 
\textsc{$\;\rightarrow$faith} & 37.31 & 15.21 & 30.12 & \textbf{31.7} & \textbf{38.8} \\ \hline \end{tabular}
\caption{\textsc{rouge} and faithfulness/factuality scores for \bencdec plus systems that use textual entailment as a criteria or fine-tuned on faithful annotations.}
\label{tab:entail-select}
\vspace{-0.5cm}
\end{table}

\subsection{Model Selection with Entailment}
\label{subsec:modelselection}

Our study suggests that entailment could be used as an automatic measure for faithfulness. However, we should point out that this measure is \emph{reference-less}.
Thus, it can easily be gamed, i.e., the first sentence of any source document is always entailed by the whole document. Because of this, entailment-based measures for evaluation need to be coupled with \emph{reference-based} measures like \textsc{rouge}.

However, one major advantage of the measure being reference-less is that we can use it as a model selection objective or during decoding. We tested the former. Specifically, we used the probability that a summary is entailed by a document as a selection criteria to select a summary between four candidates generated by systems evaluated: \ptgen, \tconv, \tencdec, and \bencdec. Results are shown in the \textsc{entail} row of Table~\ref{tab:entail-select}. We can see that indeed this is a strong metric to optimize towards if we want faithful summaries - almost 5\% absolute better. There is a trade-off in terms of \textsc{rouge}, but this model must select amongst 4 systems, 3 of which have significantly lower \textsc{rouge} than the best model.

A further experiment is to train a model explicitly to predict faithfulness. In order to do this, we further fine-tuned the entailment model  using the `faithful' annotations generated during our evaluation. For all summary-document pairs marked as `faithful', we set the associated class to `entailment', otherwise we set it to `neutral'. This allowed for us to also fine-tune the last classification layers taking advantage of the correlation between `entailment' and `faithfulness'. Results using 5-fold cross validation are shown in the \textsc{entail$\rightarrow$faith} row of Table~\ref{tab:entail-select}. We see here that indeed this does improve the ability to select faithful summaries from a set of candidates, though slightly. We would expect to see larger gains with more training data. However, this model is significantly better than \textsc{entail} on \textsc{rouge}-based metrics and seems like a good balance between \textsc{rouge} and better faithfulness.

\section{Related Work}
\label{sec:related}

Following the Document Understanding Conference \citep[DUC;][]{dang2005overview}, a majority of work has focused on evaluating the content and the linguistic quality of summaries \cite{Nenkova:2005:ATS}. Most popular among them is the automatic metric ROUGE \cite{rouge} that measures the unigram and bigram overlap (ROUGE-1 and ROUGE-2) as a proxy for assessing informativeness and the longest common subsequence (ROUGE-L), for fluency. ROUGE, however, can be misleading when used as the only means to assess the informativeness of summaries \cite{schluter:2017:EACLshort}. Hence, the ROUGE score is often complemented with subjective human assessment of summaries.
More objective measures have been proposed to improve agreement among human annotators. Pyramid method \cite{Nenkova2004a} requires summaries to be annotated by experts for salient information. \newcite{narayan18:xsum,Narayan2018Ranking} used a question-answering based approach where a summary is used as context to answer questions which were written based on its reference summary. \newcite{hardy-etal-2019-highres} proposed a reference-less approach where a summary is assessed against the source document, highlighted with its pertinent content. 

There has not been much work on evaluating faithfulness and truthfulness of abstractive summaries. The automatic evaluation such as ROUGE and the human evaluation of saliency and linguistic quality of summaries are not sufficient due to the complex nature of the task. Recently, \newcite{bansal2018fastabstractive} asked human annotators to assess the summary relevance measuring both the saliency and the presence of contradictory/unrelated information. \newcite{dhingra-etal-2019-handling} proposed a new automatic metric, PARENT, for data-to-text generation \cite{lebret2016neural,wiseman-etal-2017-challenges} which aligns $n$-grams from the reference and generated texts to the source table to measure the accuracy of $n$-grams that are entailed from the source table. \newcite{fact-brainsum} proposed a model-based automatic metric to assess the faithfulness of Wikipedia summaries;  they trained an end-to-end model to extract a complete set of OpenIE-style \cite{openie} facts from both the source text and the generated summary. The summary is faithful if it is precise in generating facts from the source text. 
In our experiments with OpenIE-based measures, we found that they are not suited for evaluating extreme summarization models; all models perform poorly on these metrics without any significant differences. Like ours, few recent works (some in parallel) have explored natural language inference and question answering models to detect factual consistency in generated text \cite{welleck-etal-2019-dialogue,falke-etal-2019-ranking,Kryscinski2019EvaluatingTF,Wang2020AskingAA}. In line with our findings, \newcite{falke-etal-2019-ranking} observed that the BERT-based NLI models substantially improved summaries reranking in terms of their correctness. \newcite{Kryscinski2019EvaluatingTF} proposed an NLI-based fact checking model that is trained on a dataset tailored for detecting factual inconsistencies in generated text. \newcite{Wang2020AskingAA} proposed a question answering and generation based automatic evaluation protocol that is designed to identify factual inconsistencies in a generated summary. Future work will likely investigate better ways of generating questions and measuring factual consistency to address poor correlation with faithfulness and factuality annotations.

Finally, others have used reinforcement learning to improve informativeness and reduce contradictory information in abstractive summaries, e.g., \newcite{Pasunuru-multireward18} used a textual entailment-based reward and \newcite{arumae-liu-2019-guiding}, a question-answering based reward. However, these approaches don't evaluate if these rewards improve faithfulness of summaries.

\section{Conclusion}

We conducted a large-scale study of hallucinations in abstractive document summarization. We found that (i) tackling hallucination is a critical challenge for abstractive summarization, perhaps the most critical, (ii) NLU-driven pretraining in neural text generators is key to generate informative, coherent, faithful and factual abstracts, but it is still far from solving the problem; and (iii) measures such as ROUGE or BERTScore will not be sufficient when studying the problem; semantic inference-based automatic measures are better representations of true summarization quality.

\section*{Acknowledgments}

We thank Ratish Puduppully, Yova Kementchedjhieva, Ankur Parikh, Peter Liu, Slav Petrov, the reviewers and the action editor for invaluable feedback. The hard work of Muqthar Mohammad, Mohd Majeed and Ashwin Kakarla made our human annotation possible.

\bibliography{hallucinations}

\begin{thebibliography}{58}
\expandafter\ifx\csname natexlab\endcsname\relax\def\natexlab#1{#1}\fi

\bibitem[{Alberti et~al.(2019)Alberti, Andor, Pitler, Devlin, and
  Collins}]{alberti-etal-2019-synthetic}
Chris Alberti, Daniel Andor, Emily Pitler, Jacob Devlin, and Michael Collins.
  2019.
\newblock Synthetic {QA} corpora generation with roundtrip consistency.
\newblock In \emph{Proceedings of the 57th Annual Meeting of the Association
  for Computational Linguistics}, pages 6168--6173, Florence, Italy.

\bibitem[{Arumae and Liu(2019)}]{arumae-liu-2019-guiding}
Kristjan Arumae and Fei Liu. 2019.
\newblock Guiding extractive summarization with question-answering rewards.
\newblock In \emph{Proceedings of the 2019 Conference of the North {A}merican
  Chapter of the Association for Computational Linguistics: Human Language
  Technologies}, pages 2566--2577, Minneapolis, Minnesota.

\bibitem[{Bahdanau et~al.(2015)Bahdanau, Cho, and Bengio}]{bahdanau_2014}
Dzmitry Bahdanau, Kyunghyun Cho, and Yoshua Bengio. 2015.
\newblock Neural machine translation by jointly learning to align and
  translate.
\newblock In \emph{3rd International Conference on Learning Representations},
  San Diego, CA, USA.

\bibitem[{Banko et~al.(2007)Banko, Cafarella, Soderland, Broadhead, and
  Etzioni}]{openie}
Michele Banko, Michael~J. Cafarella, Stephen Soderland, Matt Broadhead, and
  Oren Etzioni. 2007.
\newblock Open information extraction from the web.
\newblock In \emph{Proceedings of the 20th International Joint Conference on
  Artifical Intelligence}, pages 2670--2676, Hyderabad, India.

\bibitem[{Brown and Day(1983)}]{brown1983}
Ann~L. Brown and Jeanne~D. Day. 1983.
\newblock Macrorules for summarizing texts: {T}he development of expertise.
\newblock \emph{Journal of Verbal Learning and Verbal Behaviour}, 22(1):1--14.

\bibitem[{Chen and Bansal(2018)}]{bansal2018fastabstractive}
Yen-Chun Chen and Mohit Bansal. 2018.
\newblock Fast abstractive summarization with reinforce-selected sentence
  rewriting.
\newblock In \emph{Proceedings of the 56th Annual Meeting of the Association
  for Computational Linguistics}, pages 675--686, Melbourne, Australia.

\bibitem[{Dai et~al.(2019)Dai, Yang, Yang, Carbonell, Le, and
  Salakhutdinov}]{transformetxl}
Zihang Dai, Zhilin Yang, Yiming Yang, Jaime Carbonell, Quoc Le, and Ruslan
  Salakhutdinov. 2019.
\newblock Transformer-{XL}: Attentive language models beyond a fixed-length
  context.
\newblock In \emph{Proceedings of the 57th Annual Meeting of the Association
  for Computational Linguistics}, pages 2978--2988, Florence, Italy.

\bibitem[{Dang(2005)}]{dang2005overview}
Hoa~Trang Dang. 2005.
\newblock {Overview of DUC 2005}.
\newblock In \emph{Proceedings of the Document Understanding Conference}, pages
  1--12.

\bibitem[{Devlin et~al.(2019)Devlin, Chang, Lee, and Toutanova}]{bert}
Jacob Devlin, Ming-Wei Chang, Kenton Lee, and Kristina Toutanova. 2019.
\newblock {BERT}: Pre-training of deep bidirectional transformers for language
  understanding.
\newblock In \emph{Proceedings of the 2019 Conference of the North {A}merican
  Chapter of the Association for Computational Linguistics: Human Language
  Technologies}, pages 4171--4186, Minneapolis, Minnesota.

\bibitem[{Dhingra et~al.(2019)Dhingra, Faruqui, Parikh, Chang, Das, and
  Cohen}]{dhingra-etal-2019-handling}
Bhuwan Dhingra, Manaal Faruqui, Ankur Parikh, Ming-Wei Chang, Dipanjan Das, and
  William Cohen. 2019.
\newblock Handling divergent reference texts when evaluating table-to-text
  generation.
\newblock In \emph{Proceedings of the 57th Annual Meeting of the Association
  for Computational Linguistics}, pages 4884--4895, Florence, Italy.

\bibitem[{van Dijk and Kintsch(1978)}]{dijk1978}
Teun~A. van Dijk and Walter Kintsch. 1978.
\newblock Cognitive psychology and discourse: Recalling and summarizing
  stories.
\newblock In Wolfgang~U. Dressler, editor, \emph{Current Trends in
  Textlinguistics}, pages 61--80.

\bibitem[{Dong et~al.(2019)Dong, Yang, Wang, Wei, Liu, Wang, Gao, Zhou, and
  Hon}]{unilm_arxiv19}
Li~Dong, Nan Yang, Wenhui Wang, Furu Wei, Xiaodong Liu, Yu~Wang, Jianfeng Gao,
  Ming Zhou, and Hsiao-Wuen Hon. 2019.
\newblock Unified language model pre-training for natural language
  understanding and generation.
\newblock In \emph{Advances in Neural Information Processing Systems 32}, pages
  13063--13075. Curran Associates, Inc.

\bibitem[{Falke et~al.(2019)Falke, Ribeiro, Utama, Dagan, and
  Gurevych}]{falke-etal-2019-ranking}
Tobias Falke, Leonardo F.~R. Ribeiro, Prasetya~Ajie Utama, Ido Dagan, and Iryna
  Gurevych. 2019.
\newblock Ranking generated summaries by correctness: An interesting but
  challenging application for natural language inference.
\newblock In \emph{Proceedings of the 57th Annual Meeting of the Association
  for Computational Linguistics}, pages 2214--2220, Florence, Italy.

\bibitem[{Gehring et~al.(2017)Gehring, Auli, Grangier, Yarats, and
  Dauphin}]{convseq2seq}
Jonas Gehring, Michael Auli, David Grangier, Denis Yarats, and Yann~N. Dauphin.
  2017.
\newblock Convolutional sequence to sequence learning.
\newblock In \emph{Proceedings of the 34th International Conference on Machine
  Learning}, volume~70, pages 1243–--1252, Sydney, NSW, Australia.

\bibitem[{Gehrmann et~al.(2018)Gehrmann, Deng, and Rush}]{gehrmann2018bottom}
Sebastian Gehrmann, Yuntian Deng, and Alexander Rush. 2018.
\newblock Bottom-up abstractive summarization.
\newblock In \emph{Proceedings of the 2018 Conference on Empirical Methods in
  Natural Language Processing}, pages 4098--4109, Brussels, Belgium.

\bibitem[{Goodrich et~al.(2019)Goodrich, Rao, Liu, and Saleh}]{fact-brainsum}
Ben Goodrich, Vinay Rao, Peter~J. Liu, and Mohammad Saleh. 2019.
\newblock Assessing the factual accuracy of generated text.
\newblock In \emph{Proceedings of the 25th ACM SIGKDD International Conference
  on Knowledge Discovery and Data Mining}, pages 166--175, New York, NY, USA.

\bibitem[{Gu et~al.(2016)Gu, Lu, Li, and Li}]{Gu2016Incorporating}
Jiatao Gu, Zhengdong Lu, Hang Li, and Victor~O.K. Li. 2016.
\newblock Incorporating copying mechanism in sequence-to-sequence learning.
\newblock In \emph{Proceedings of the 54th Annual Meeting of the Association
  for Computational Linguistics}, pages 1631--1640, Berlin, Germany.

\bibitem[{Hardy et~al.(2019)Hardy, Narayan, and
  Vlachos}]{hardy-etal-2019-highres}
Hardy, Shashi Narayan, and Andreas Vlachos. 2019.
\newblock {H}igh{RES}: Highlight-based reference-less evaluation of
  summarization.
\newblock In \emph{Proceedings of the 57th Annual Meeting of the Association
  for Computational Linguistics}, pages 3381--3392, Florence, Italy.

\bibitem[{Hermann et~al.(2015)Hermann, Ko\v{c}isk\'{y}, Grefenstette, Espeholt,
  Kay, Suleyman, and Blunsom}]{hermann-nips15}
Karl~Moritz Hermann, Tom\'{a}\v{s} Ko\v{c}isk\'{y}, Edward Grefenstette, Lasse
  Espeholt, Will Kay, Mustafa Suleyman, and Phil Blunsom. 2015.
\newblock Teaching machines to read and comprehend.
\newblock In \emph{Advances in Neural Information Processing Systems 28}, pages
  1693--1701. Curran Associates, Inc.

\bibitem[{Holtzman et~al.(2020)Holtzman, Buys, Forbes, and
  Choi}]{holtzman-arxiv19}
Ari Holtzman, Jan Buys, Maxwell Forbes, and Yejin Choi. 2020.
\newblock The curious case of neural text degeneration.
\newblock In \emph{Proceedings of the 8th International Conference on Learning
  Representations}, Virtual Conference, Formerly Addis Ababa Ethiopia.

\bibitem[{Khandelwal et~al.(2019)Khandelwal, Clark, Jurafsky, and
  Kaiser}]{khandelwal_2019}
Urvashi Khandelwal, Kevin Clark, Dan Jurafsky, and Lukasz Kaiser. 2019.
\newblock Sample efficient text summarization using a single pre-trained
  transformer.
\newblock \emph{CoRR}, abs/1905.08836.

\bibitem[{Kryscinski et~al.(2019{\natexlab{a}})Kryscinski, Keskar, McCann,
  Xiong, and Socher}]{Kryscinski2019NeuralTS}
Wojciech Kryscinski, Nitish~Shirish Keskar, Bryan McCann, Caiming Xiong, and
  Richard Socher. 2019{\natexlab{a}}.
\newblock Neural text summarization: A critical evaluation.
\newblock In \emph{Proceedings of the 2019 Conference on Empirical Methods in
  Natural Language Processing and the 9th International Joint Conference on
  Natural Language Processing}, pages 540--551, Hong Kong, China.

\bibitem[{Kryscinski et~al.(2019{\natexlab{b}})Kryscinski, McCann, Xiong, and
  Socher}]{Kryscinski2019EvaluatingTF}
Wojciech Kryscinski, Bryan McCann, Caiming Xiong, and Richard Socher.
  2019{\natexlab{b}}.
\newblock Evaluating the factual consistency of abstractive text summarization.
\newblock \emph{CoRR}, abs/1910.12840.

\bibitem[{Kudo and Richardson(2018)}]{sentencepiece}
Taku Kudo and John Richardson. 2018.
\newblock Sentencepiece: {A} simple and language independent subword tokenizer
  and detokenizer for neural text processing.
\newblock \emph{CoRR}, abs/1808.06226.

\bibitem[{Kwiatkowski et~al.(2019)Kwiatkowski, Palomaki, Redfield, Collins,
  Parikh, Alberti, Epstein, Polosukhin, Devlin, Lee, Toutanova, Jones, Kelcey,
  Chang, Dai, Uszkoreit, Le, and Petrov}]{nq}
Tom Kwiatkowski, Jennimaria Palomaki, Olivia Redfield, Michael Collins, Ankur
  Parikh, Chris Alberti, Danielle Epstein, Illia Polosukhin, Jacob Devlin,
  Kenton Lee, Kristina Toutanova, Llion Jones, Matthew Kelcey, Ming-Wei Chang,
  Andrew~M. Dai, Jakob Uszkoreit, Quoc Le, and Slav Petrov. 2019.
\newblock Natural questions: A benchmark for question answering research.
\newblock \emph{Transactions of the Association for Computational Linguistics},
  7:453--466.

\bibitem[{Landis and Koch(1977)}]{landis77}
J.~Richard Landis and Gary~G. Koch. 1977.
\newblock The measurement of observer agreement for categorical data.
\newblock \emph{Biometrics}, 33(1):159--174.

\bibitem[{Lebret et~al.(2016)Lebret, Grangier, and Auli}]{lebret2016neural}
R{\'e}mi Lebret, David Grangier, and Michael Auli. 2016.
\newblock Neural text generation from structured data with application to the
  biography domain.
\newblock In \emph{Proceedings of the 2016 Conference on Empirical Methods in
  Natural Language Processing}, pages 1203--1213, Austin, Texas.

\bibitem[{Lin and Hovy(2003)}]{rouge}
Chin~Yew Lin and Eduard Hovy. 2003.
\newblock Automatic evaluation of summaries using n-gram co-occurrence
  statistics.
\newblock In \emph{Proceedings of the 2003 Human Language Technology Conference
  of the North American Chapter of the Association for Computational
  Linguistics}, pages 150--157.

\bibitem[{Liu et~al.(2018)Liu, Saleh, Pot, Goodrich, Sepassi, Kaiser, and
  Shazeer}]{wikisum}
Peter~J. Liu, Mohammad Saleh, Etienne Pot, Ben Goodrich, Ryan Sepassi, Lukasz
  Kaiser, and Noam Shazeer. 2018.
\newblock Generating wikipedia by summarizing long sequences.
\newblock In \emph{Proceedings of the 6th International Conference on Learning
  Representations}, Vancouver Canada.

\bibitem[{Liu et~al.(2019)Liu, Ott, Goyal, Du, Joshi, Chen, Levy, Lewis,
  Zettlemoyer, and Stoyanov}]{roberta}
Yinhan Liu, Myle Ott, Naman Goyal, Jingfei Du, Mandar Joshi, Danqi Chen, Omer
  Levy, Mike Lewis, Luke Zettlemoyer, and Veselin Stoyanov. 2019.
\newblock {RoBERTa: A} robustly optimized {BERT} pretraining approach.
\newblock \emph{CoRR}, abs/1907.11692.

\bibitem[{Mani(2001)}]{mani2001automatic}
Inderjeet Mani. 2001.
\newblock \emph{Automatic summarization}, volume~3.
\newblock John Benjamins Publishing.

\bibitem[{Narayan et~al.(2018{\natexlab{a}})Narayan, Cohen, and
  Lapata}]{narayan18:xsum}
Shashi Narayan, Shay~B. Cohen, and Mirella Lapata. 2018{\natexlab{a}}.
\newblock Don't give me the details, just the summary! {T}opic-aware
  convolutional neural networks for extreme summarization.
\newblock In \emph{Proceedings of the 2018 Conference on Empirical Methods in
  Natural Language Processing}, pages 1797--1807, Brussels, Belgium.

\bibitem[{Narayan et~al.(2018{\natexlab{b}})Narayan, Cohen, and
  Lapata}]{Narayan2018Ranking}
Shashi Narayan, Shay~B. Cohen, and Mirella Lapata. 2018{\natexlab{b}}.
\newblock Ranking sentences for extractive summarization with reinforcement
  learning.
\newblock In \emph{Proceedings of the 2018 Conference of the North American
  Chapter of the Association for Computational Linguistics: Human Language
  Technologies}, pages 1747--1759, New Orleans, Louisiana.

\bibitem[{Narayan et~al.(2020)Narayan, Simoes, Ma, Craighead, and
  McDonald}]{Narayan2020QURIOUSQG}
Shashi Narayan, Gon\c{c}alo Simoes, Ji~Ma, Hannah Craighead, and Ryan~T.
  McDonald. 2020.
\newblock {QURIOUS: Q}uestion generation pretraining for text generation.
\newblock \emph{CoRR}, abs/2004.11026.

\bibitem[{Nenkova(2005)}]{Nenkova:2005:ATS}
Ani Nenkova. 2005.
\newblock {Automatic Text Summarization of Newswire: Lessons Learned from the
  Document Understanding Conference}.
\newblock In \emph{Proceedings of the 20th National Conference on Artificial
  Intelligence - Volume 3}, pages 1436--1441.

\bibitem[{Nenkova and McKeown(2011)}]{Nenkova:McKeown:2011}
Ani Nenkova and Kathleen McKeown. 2011.
\newblock Automatic summarization.
\newblock \emph{Foundations and Trends in Information Retrieval},
  5(2--3):103--233.

\bibitem[{Nenkova and Passonneau(2004)}]{Nenkova2004a}
Ani Nenkova and Rebecca Passonneau. 2004.
\newblock Evaluating content selection in summarization: {The Pyramid} method.
\newblock In \emph{Proceedings of the Human Language Technology Conference of
  the North {A}merican Chapter of the Association for Computational
  Linguistics}, pages 145--152, Boston, Massachusetts, USA.

\bibitem[{Niven and Kao(2019)}]{niven2019}
Timothy Niven and Hung-Yu Kao. 2019.
\newblock Probing neural network comprehension of natural language arguments.
\newblock In \emph{Proceedings of the 57th Annual Meeting of the Association
  for Computational Linguistics}, pages 4658--4664, Florence, Italy.

\bibitem[{Pasunuru and Bansal(2018)}]{Pasunuru-multireward18}
Ramakanth Pasunuru and Mohit Bansal. 2018.
\newblock Multi-reward reinforced summarization with saliency and entailment.
\newblock In \emph{Proceedings of the 2018 Conference of the North American
  Chapter of the Association for Computational Linguistics: Human Language
  Technologies}, pages 646--653, New Orleans, Louisiana.

\bibitem[{Paulus et~al.(2018)Paulus, Xiong, and Socher}]{Paulus2018Deep}
Romain Paulus, Caiming Xiong, and Richard Socher. 2018.
\newblock A deep reinforced model for abstractive summarization.
\newblock In \emph{Proceedings of the 6th International Conference on Learning
  Representations}, Vancouver, BC, Canada.

\bibitem[{Radford et~al.(2018)Radford, Narasimhan, Salimans, and
  Sutskever}]{gpt}
Alec Radford, Karthik Narasimhan, Tim Salimans, and Ilya Sutskever. 2018.
\newblock Improving language understanding by generative pre-training.
\newblock Technical report.

\bibitem[{Radford et~al.(2019)Radford, Wu, Child, Luan, Amodei, and
  Sutskever}]{gpt2}
Alec Radford, Jeff Wu, Rewon Child, David Luan, Dario Amodei, and Ilya
  Sutskever. 2019.
\newblock Language models are unsupervised multitask learners.
\newblock Technical report.

\bibitem[{Rothe et~al.(2020)Rothe, Narayan, and Severyn}]{berts2s}
Sascha Rothe, Shashi Narayan, and Aliaksei Severyn. 2020.
\newblock Leveraging pre-trained checkpoints for sequence generation tasks.
\newblock \emph{To appear in Transactions of the Association for Computational
  Linguistics}, abs/1907.12461.

\bibitem[{Sandhaus(2008)}]{nytcorpus}
Evan Sandhaus. 2008.
\newblock {The New York Times Annotated Corpus}.
\newblock \emph{Linguistic Data Consortium, Philadelphia}, 6(12).

\bibitem[{Schluter(2017)}]{schluter:2017:EACLshort}
Natalie Schluter. 2017.
\newblock The limits of automatic summarisation according to rouge.
\newblock In \emph{Proceedings of the 15th Conference of the European Chapter
  of the Association for Computational Linguistics}, pages 41--45, Valencia,
  Spain.

\bibitem[{See et~al.(2017)See, Liu, and Manning}]{see-acl17}
Abigail See, Peter~J. Liu, and Christopher~D. Manning. 2017.
\newblock Get to the point: {S}ummarization with pointer-generator networks.
\newblock In \emph{Proceedings of the 55th Annual Meeting of the Association
  for Computational Linguistics}, pages 1073--1083, Vancouver, Canada.

\bibitem[{See et~al.(2019)See, Pappu, Saxena, Yerukola, and
  Manning}]{see2019massively}
Abigail See, Aneesh Pappu, Rohun Saxena, Akhila Yerukola, and Christopher~D.
  Manning. 2019.
\newblock Do massively pretrained language models make better storytellers?
\newblock In \emph{Proceedings of the 23rd Conference on Computational Natural
  Language Learning}, pages 843--861, Hong Kong, China.

\bibitem[{Song et~al.(2019)Song, Tan, Qin, Lu, and Liu}]{mass_icml19}
Kaitao Song, Xu~Tan, Tao Qin, Jianfeng Lu, and Tie{-}Yan Liu. 2019.
\newblock {MASS: M}asked sequence to sequence pre-training for language
  generation.
\newblock In \emph{Proceedings of the 36th International Conference on Machine
  Learning}, Long Beach, California.

\bibitem[{Sutskever et~al.(2014)Sutskever, Vinyals, and
  Le}]{Sutskever2014Sequence}
Ilya Sutskever, Oriol Vinyals, and Quoc~V Le. 2014.
\newblock Sequence to sequence learning with neural networks.
\newblock In \emph{Advances in Neural Information Processing Systems 27}, pages
  3104--3112. Curran Associates, Inc.

\bibitem[{Vaswani et~al.(2017)Vaswani, Shazeer, Parmar, Uszkoreit, Jones,
  Gomez, Kaiser, and Polosukhin}]{transformer}
Ashish Vaswani, Noam Shazeer, Niki Parmar, Jakob Uszkoreit, Llion Jones,
  Aidan~N Gomez, Lukasz Kaiser, and Illia Polosukhin. 2017.
\newblock Attention is all you need.
\newblock In \emph{Advances in Neural Information Processing Systems 30}, pages
  5998--6008. Curran Associates, Inc.

\bibitem[{Wang et~al.(2020)Wang, Cho, and Lewis}]{Wang2020AskingAA}
Alex Wang, Kyunghyun Cho, and Michael Lewis. 2020.
\newblock Asking and answering questions to evaluate the factual consistency of
  summaries.
\newblock In \emph{Proceedings of the 58th Annual Meeting of the Association
  for Computational Linguistics}, Virtual Conference, Formerly Seattle, USA.

\bibitem[{Welleck et~al.(2020)Welleck, Kulikov, Roller, Dinan, Cho, and
  Weston}]{welleck-arxiv19}
Sean Welleck, Ilia Kulikov, Stephen Roller, Emily Dinan, Kyunghyun Cho, and
  Jason Weston. 2020.
\newblock Neural text generation with unlikelihood training.
\newblock In \emph{Proceedings of the 8th International Conference on Learning
  Representations}, Virtual Conference, Formerly Addis Ababa Ethiopia.

\bibitem[{Welleck et~al.(2019)Welleck, Weston, Szlam, and
  Cho}]{welleck-etal-2019-dialogue}
Sean Welleck, Jason Weston, Arthur Szlam, and Kyunghyun Cho. 2019.
\newblock Dialogue natural language inference.
\newblock In \emph{Proceedings of the 57th Annual Meeting of the Association
  for Computational Linguistics}, pages 3731--3741, Florence, Italy.

\bibitem[{Williams et~al.(2018)Williams, Nangia, and Bowman}]{mnli}
Adina Williams, Nikita Nangia, and Samuel Bowman. 2018.
\newblock A broad-coverage challenge corpus for sentence understanding through
  inference.
\newblock In \emph{Proceedings of the 2018 Conference of the North American
  Chapter of the Association for Computational Linguistics: Human Language
  Technologies}, pages 1112--1122, New Orleans, Louisiana.

\bibitem[{Wiseman et~al.(2017)Wiseman, Shieber, and
  Rush}]{wiseman-etal-2017-challenges}
Sam Wiseman, Stuart Shieber, and Alexander Rush. 2017.
\newblock Challenges in data-to-document generation.
\newblock In \emph{Proceedings of the 2017 Conference on Empirical Methods in
  Natural Language Processing}, pages 2253--2263, Copenhagen, Denmark.

\bibitem[{Wu et~al.(2016)Wu, Schuster, Chen, Le, Norouzi, Macherey, Krikun,
  Cao, Gao, Macherey, Klingner, Shah, Johnson, Liu, Kaiser, Gouws, Kato, Kudo,
  Kazawa, Stevens, Kurian, Patil, Wang, Young, Smith, Riesa, Rudnick, Vinyals,
  Corrado, Hughes, and Dean}]{wordpiece}
Yonghui Wu, Mike Schuster, Zhifeng Chen, Quoc~V. Le, Mohammad Norouzi, Wolfgang
  Macherey, Maxim Krikun, Yuan Cao, Qin Gao, Klaus Macherey, Jeff Klingner,
  Apurva Shah, Melvin Johnson, Xiaobing Liu, Lukasz Kaiser, Stephan Gouws,
  Yoshikiyo Kato, Taku Kudo, Hideto Kazawa, Keith Stevens, George Kurian,
  Nishant Patil, Wei Wang, Cliff Young, Jason Smith, Jason Riesa, Alex Rudnick,
  Oriol Vinyals, Greg Corrado, Macduff Hughes, and Jeffrey Dean. 2016.
\newblock Google's neural machine translation system: Bridging the gap between
  human and machine translation.
\newblock \emph{CoRR}, abs/1609.08144.

\bibitem[{Yang et~al.(2019)Yang, Dai, Yang, Carbonell, Salakhutdinov, and
  Le}]{xlnet_arxiv19}
Zhilin Yang, Zihang Dai, Yiming Yang, Jaime~G. Carbonell, Ruslan Salakhutdinov,
  and Quoc~V. Le. 2019.
\newblock {XLNet: G}eneralized autoregressive pretraining for language
  understanding.
\newblock \emph{CoRR}, abs/1906.08237.

\bibitem[{Zhang et~al.(2020)Zhang, Kishore, Wu, Weinberger, and
  Artzi}]{bertscore}
Tianyi Zhang, Varsha Kishore, Felix Wu, Kilian~Q. Weinberger, and Yoav Artzi.
  2020.
\newblock {BERTScore: E}valuating text generation with {BERT}.
\newblock In \emph{Proceedings of the 8th International Conference on Learning
  Representations}, Virtual Conference, Formerly Addis Ababa Ethiopia.

\end{thebibliography}
\bibliographystyle{acl_natbib}

\clearpage

\appendix

\section{Model Hyperparameters and Predictions} 

\ptgen and \tconv model predictions are provided by \newcite{narayan18:xsum} and Transformer model predictions from \gpttune, \tencdec and \bencdec, by \newcite{berts2s}. 
Both \ptgen and \tconv use a Stanford tokenized vocabulary size of 50k. \tencdec and \bencdec use a vocabulary size of around $\sim$30k WordPieces \cite{wordpiece} to match BERT pretrained vocabulary and, 
\gpttune, a vocabulary size of around $\sim$50k SentencePieces \cite{sentencepiece} to match the GPT-2 pretrained vocabulary. All models use the same uncased vocabulary on both source and target sides. Both \ptgen and \tconv summaries were generated using beam search with beam size 10, the Transformer models use beam size of 4. See \newcite{narayan18:xsum} and \newcite{berts2s} for more details on these models. 

\begin{table}[ht!]
\centering
\footnotesize
\begin{tabular}{l|cccc}
\hline
\multirow{2}{*}{\textbf{Models}} & \multicolumn{4}{|c}{\textbf{Fleiss' Kappa}} \\
& \textbf{Hall.} & \textbf{Fact.} & \textbf{Rept.} & \textbf{Inco.}\\ \hline
\ptgen & 0.70 & 0.91 & 0.89 & 0.84 \\ 
\tconv & 0.73 & 0.91 & 0.93 & 0.90 \\
\tencdec & 0.67 & 0.91 & 0.92 & 0.90 \\
\bencdec & 0.67 & 0.88 & 0.94 & 0.93 \\
\gold & 0.71 & --- & 1.00 & 0.98\\ \hline
\end{tabular}
\caption{Fleiss's Kappa scores measuring word-level agreements among annotators for different annotation tasks: hallucination (Hall.), factuality (Fact.), repetition (Rept.) and incoherence (Inco.) assessments.}
\label{app-tab:kappa}
\vspace{-0.1cm}
\end{table}

\section{Inter annotator agreement}

We  estimated Fleiss's Kappa ($k$) to assess the agreement among our raters when categorizing a word in the summary as one of faithful, intrinsically hallucinated and extrinsically hallucinated. The results are shown in Table~\ref{app-tab:kappa}. All models showed substantial agreement \citep[0.61 $\leq k \leq $ 0.80;][]{landis77} among their annotations. 

Table~\ref{app-tab:kappa} also shows Fleiss's Kappa ($k$) to assess the agreement among our raters for factuality. All models showed almost perfect agreement \citep[0.81 $\leq k \leq $ 1.0;][]{landis77} among their annotations. 

\begin{table}[ht!]
\centering
\footnotesize
\begin{tabular}{l | c | c | c}
\hline
\multirow{2}{*}{\textbf{Models}} & \textbf{Intrinsic} & \textbf{Extrinsic} & \multirow{2}{*}{\parbox[t]{0.8cm}{\textbf{avg.\\length}}} \\ 
& \textbf{total (avg.)} & \textbf{total (avg.)} &  \\ \hline
\ptgen & 625 (1.35) & \textbf{1424 (2.85)} & 8.48 \\ 
\tconv & 518 (1.04) & 1556 (3.11) & 8.44 \\
\tencdec & 589 (1.18) & 1556 (3.11) & 7.39 \\
\bencdec & 530 (1.06)  & 1520 (3.04) & \textbf{6.12} \\
\gold & \textbf{276 (0.55)} & 1807 (3.61) & 7.11 \\ \hline
\end{tabular}
\caption{Total number of spans and the average number of spans per document, annotated as intrinsic or extrinsic hallucinations for all 500 document-summary pairs by three annotators. We also show the average span length for each system.}
\label{app-tab:intexhal-1}
\vspace{-0.1cm}
\end{table}

\begin{table}[ht!]
\centering
\footnotesize
\begin{tabular}{l|c|c}
\hline
\textbf{Models} & \textbf{Repetition} & \textbf{Incoherence} \\ \hline
\ptgen & 17.5 & 20.3  \\ 
\tconv & 16.7 & 17.7  \\
\tencdec & 8.9 & 11.5 \\
\bencdec & 8.7 & 9.5 \\
\gold & \textbf{0.0} & \textbf{0.8} \\ \hline
\end{tabular}
\caption{Repetition and Incoherence Evaluation. The numbers show the the percentage of 500 summaries where at least one word in a summary was annotated by all three annotators with the ``Repetition'' or ``Incoherence'' related issue. The lowest numbers are boldfaced.}
\label{app-tab:repincoh}
\vspace{-0.1cm}
\end{table}

\begin{table}[ht!]
\centering
\footnotesize
\begin{tabular}{l | c  c}
\hline
\textbf{Metric} & \textbf{Faithful} & \textbf{Factual} \\ \hline
\textsc{rouge-1} & 0.197 & 0.125 \\
\textsc{rouge-2} & 0.162 & 0.095 \\
\textsc{rouge-l} & 0.162 & 0.113 \\
BERTScore & 0.190 & 0.116 \\ \hline
Repetition & 0.064 & 0.075 \\
Incoherence & 0.067 & 0.082 \\ \hline
QA & 0.044 & 0.027 \\ 
Entailment & \textbf{0.431} & \textbf{0.264} \\
\hline
\end{tabular}
\caption{Spearman's correlation coefficient ($|r_s|$) of different metrics with faithful and factual annotations.}
\label{app-tab:correl}
\vspace{-0.1cm}
\end{table}

\begin{table*}[ht!]
\centering
\footnotesize
\begin{tabular}{l|c|cc|cc|cc|c}
\hline
\multirow{3}{*}{\textbf{Models}} & \multirow{3}{*}{\textbf{Faithful}} & \multicolumn{6}{|c|}{\textbf{Hallucinated}} &  \multirow{3}{*}{\textbf{Factual}}\\
& & \multicolumn{2}{|c}{\textbf{I}} & \multicolumn{2}{|c}{\textbf{E}}  & \multicolumn{2}{|c|}{\textbf{I $\cup$ E}}  & \\
& & \textbf{total} & \textbf{factual} & \textbf{total} & \textbf{factual} & \textbf{total} & \textbf{factual} & \\ \hline
\ptgen & 24.7 & 19.9 & 0.4 & \textbf{63.3} & 2.2 & 75.3 & 2.6 & 27.3  \\ 
\tconv & 21.5 & 17.7 & 0.8 & 71.5 & 5.0 & 78.5  & 5.4 & 26.9  \\
\tencdec & 20.7 & 19.1 & 1.4 & 68.1 & 3.4 & 79.3 & 4.6 & 25.3 \\
\bencdec & \textbf{26.9} & 16.9 & \textbf{1.8} & 64.1 & \textbf{6.6} & \textbf{73.1} & \textbf{7.8} & \textbf{34.7}\\
\gold & 23.1 & \textbf{7.4} & --- & 73.1 & --- & 76.9 & --- & --- \\ \hline
\end{tabular}
\caption{Intrinsic vs Extrinsic Hallucinations and their factuality. The numbers in ``Hallucinated'' columns show the percentage of summaries out of 500 where at least one word was annotated by all three annotators as an intrinsic (I) or extrinsic (E) hallucination.
When a summary is not marked with any hallucination, it is ``faithful'' (1- I$\cup$E). The ``factual'' columns within the ``Hallucinated'' column show for each type (I, E and I$\cup$E), the percentage of summaries out of 500 annotated by all three annotators as factual. The final ``Factual'' column shows the total percentage of factual summaries (Faithful + I$\cup$E$_{\mbox{factual}}$). The highest numbers for faithful and factual, and the lowest numbers for hallucinations are boldfaced.}
\label{app-tab:intexhal}
\vspace{-0.1cm}
\end{table*}

\section{Highlighted Span Characteristics}

Results in Table~\ref{app-tab:intexhal-1} shed some light on the characteristics of hallucinated spans observed in different abstracts. \gold abstracts showed the least number of intrinsically hallucinated spans (0.55 per document), whereas, \ptgen abstracts showed the least number of extrinsically hallucinated spans (2.85 per document). Interestingly, the average span length for \ptgen summaries was 8.48 words, much higher than 6.12 words for \bencdec summaries. Our result demonstrates that (i) the effect of  hallucination in \bencdec is more local than what we observe in \ptgen and (ii) despite a lower number of extrinsically hallucinated spans or documents in \ptgen compared to that in \bencdec (2.85 vs 3.04 spans per document, 63.3\% vs 64.1\% documents), the total number of words that were annotated as extrinsic hallucination is much higher in \ptgen than in \bencdec (12075 vs 9302 words).

\section{Assessment of Linguistic Irregularities.}
\label{sec:humaneval-c}

Following standard practice in summarization, all 2500 document-summary pairs were annotated for repetition and incoherence related linguistic irregularities. Annotators were presented only a single-sentence summary and were asked to identify all spans of text in the summary that were either repeated or made the summary incoherent. We again elicited judgments from three different annotators for each document-summary pair. Results are shown in Table~\ref{app-tab:repincoh}.

Overall, all neural text generation systems are getting better in generating repetition-free and coherent single-sentence summaries of news articles. Transformer-based models, \tencdec and \bencdec in particular, perform superior to RNN-based \ptgen and CNN-based \tconv models. Nonetheless, Table~\ref{app-tab:correl}
shows that these metrics fail to correlate with faithful, hallucinated and factual assessments of summaries. Fleiss's Kappa ($k$) values for repetition and incoherence assessments showed almost a perfect agreement \citep[0.81 $\leq k \leq $ 1.0;][]{landis77} among our raters (see Table~\ref{app-tab:kappa}).

\section{Full Hallucination Results}

Table~\ref{app-tab:intexhal} has the full results from our human study of hallucinations.

\end{document}